\documentclass[sigconf, authorversion, nonacm]{acmart}

\usepackage{color}
\usepackage{xcolor}

\usepackage[most]{tcolorbox}
\usepackage[normalem]{ulem}

\usepackage{caption}
\usepackage{subcaption}

\usepackage{tikz}
\usetikzlibrary{positioning}

\AtBeginDocument{%
  \providecommand\BibTeX{{%
    \normalfont B\kern-0.5em{\scshape i\kern-0.25em b}\kern-0.8em\TeX}}}

\begin{document}

\title{Cheating off your neighbors:\\Improving activity recognition through corroboration}

\author{Haoxiang Yu}
\email{hxyu@utexas.edu}
\orcid{0000-0002-3518-946X}
\affiliation{%
 \institution{University of Texas at Austin}
 \city{Austin}
 \state{Texas}
 \country{USA}
}

\author{Jingyi An}
\email{jingyi.an.98@gmail.com}
\affiliation{%
 \institution{Independent Researcher}
 \country{USA}
}

\author{Evan King}
\email{e.king@utexas.edu}
\affiliation{%
 \institution{University of Texas at Austin}
 \city{Austin}
 \state{Texas}
 \country{USA}
}

\author{Edison Thomaz}
\email{ethomaz@utexas.edu}
\affiliation{%
 \institution{University of Texas at Austin}
 \city{Austin}
 \state{Texas}
 \country{USA}
}

\author{Christine Julien}
\email{c.julien@utexas.edu}
\affiliation{%
  \institution{University of Texas at Austin}
  \city{Austin}
  \state{Texas}
  \country{USA}
}

\settopmatter{printacmref=false}

\begin{abstract}

Understanding the complexity of human activities solely through an individual's data can be challenging. However, in many situations, surrounding individuals are likely performing similar activities, while existing human activity recognition approaches focus almost exclusively on individual measurements and largely ignore the {\em context} of the activity. Consider two activities: attending a small group meeting and working at an office desk. From solely an individual's perspective, it can be difficult to differentiate between these activities as they may appear very similar, even though they are markedly different.
Yet, by observing others nearby, it can be possible to distinguish between these activities. 
In this paper, we propose an approach to enhance the prediction accuracy of an individual's activities by incorporating insights from surrounding individuals. We have collected a real-world dataset from 20 participants with over 58 hours of data including activities such as attending lectures, having meetings, working in the office, and eating together.
Compared to observing a single person in isolation, our proposed approach significantly improves accuracy. We regard this work as a first step in collaborative activity recognition, opening new possibilities for understanding human activity in group settings.
\end{abstract}



\maketitle

\section{Introduction}
\label{sec:intro}

Consider a scenario involving activity recognition where a subject is seated at a table. If we base our understanding solely on this isolated observation, it would be easy to assume that the subject is merely working independently. Alternatively, widening our perspective to include others in the environment could reveal that the subject is actually part of a meeting, rather than simply working alone.
In the broad field of human activity recognition, it can be difficult to obtain a comprehensive understanding of an individual's activity when examining that individual's actions in isolation.

It is common for individuals in proximity to each other to be engaged in the same or similar activities. Examples include working in a shared office, participating in a fitness class, or attending a meeting. These situations are the focus of this work. Of course, in some cases, individuals may be engaged in entirely different tasks while happen to be close to each other. An example is a person jogging past someone eating at a sidewalk cafe. Such situations are less common and often ephemeral; they are outside our scope. Finally, sometimes individuals are performing different activities, yet contributing to a common {\em group} activity. Consider the participants in a team sport. At any given point,
some players may be running, others throwing a ball, and some standing on the sidelines. In these cases, the presence of a particular set of activities indicates a broader {\em situation} that could be recognized. Such scenarios are an expansion of the work in this paper; we discuss this direction in more detail in Section~\ref{sec:discussion}.

We design a mechanism for individual activity recognition that relies on corroborating evidence from the surroundings. For simple activity classification tasks, 
this approach can significantly improve the accuracy and confidence of a result. 
For instance, working alone in a shared environment or participating in a small group meeting may result in similar IMU data for a given individual. However, if all individuals in the vicinity of an individual are participating in a small group meeting, the participant is more likely in a group meeting than work in an office. %
The robustness of this system originates from its probabilistic framework, where potential inaccuracies or errors in individual activity recognition are mitigated by neighbors. This allows for smoothing out singular discrepancies and reinforcing the accuracy and reliability of the process.%

To design an activity recognition approach based on {\em corroboration}, we start by using a state of the art activity recognition technique applied to an individual's activity data in isolation. Then, using efficient proximity-based communication, an individual's device corroborates its classification by comparing it to the classification determined by other nearby users' activity recognition algorithms. If others nearby have reached the same conclusion, the local confidence in the classification is bolstered. The most obvious analogy is one of schoolchildren ``cheating'' of their neighbors' papers on a multiple choice test -- if a student believes the answer to a question is the first option, and, upon checking with other students sitting nearby, the first option appears to be the most popular choice, the likelihood that the choice is correct is increased.

In this context, the novel contributions of this work are:
\begin{itemize}
    \item We design a corroboration-based prediction architecture that can utilize group activity data for activity recognition. This framework is built on two significant features:
    \begin{itemize}
        \item a device-to-device dissemination structure to facilitate sharing individual activity recognition outcomes with other nearby devices and
        \item a decentralized information ensemble method that utilizes nearby information to enhance recognition results.
    \end{itemize}
    \item We collect an Inertial Measurement Unit (IMU) dataset that captures data from multiple participants concurrently engaged in the same activity. Our dataset offers insights into group activities and establishes a benchmark for the synthetic construction of datasets that model group activities.
    \item We show through evaluation that, compared to observing a single person in isolation, our proposed corroborative architecture significantly improves accuracy. %
\end{itemize}

We focus on activities that may be difficult for state-of-the-art activity recognition algorithms to reliably differentiate from one another (e.g., working alone in a shared environment vs. participating in a small group meeting vs. attending a lecture).
We also show how this corroboration-based approach can serve as a building block for expressive approaches to recognizing more abstract {\em group activity situations}, in which individuals have diverse activities that contribute to determining the overall activity of the group.

\section{Related Work}
Wearable devices like smartwatches and smart rings are increasingly utilized to record daily activities. The driving force behind these devices is the Inertial Measurement Unit (IMU), owing to its affordability, low power consumption, and small size as a sensor.
Table~\ref{tab:dataset} shows several human activity datasets collected by IMUs that have been developed and used by the research community.

Human activity recognition (HAR) research has increasingly adopted deep learning. Ord\'{o}\~{n}ez and Roggen proposed a deep learning model called DeepConvLSTM,
which utilizes both convolutional and LSTM recurrent units to improve the accuracy of HAR~\cite{s16010115}. When evaluated on the OPPORTUNITY dataset~\cite{Opportunity}, DeepConvLSTM  outperformed previous non-deep learning methods.
Balli et al. conducted a comprehensive study using sensor data collected from smartwatches, leveraging traditional machine learning methodologies~\cite{balli2019human}. They gathered data from five participants engaged in eight distinct activities, experimenting with several classification algorithms. 
Of all the tested methods,
the random forest classifier outperformed the others, achieving an  accuracy of 98.1\%~\cite{balli2019human}. For traditional machine learning methods, parallel studies employing similar methodologies and alternative datasets have likewise reported comparably high levels of accuracy~\cite{9010115,8546311,THAKUR2022103417}.

The research community has also demonstrated an interest in {\em group activity recognition}~(GAR), i.e., recognizing a joint task performed by several individuals together. In contrast to the methods used for individual HAR, GAR approaches have relied predominantly on computer vision applied to video data~\cite{10.5555/3061053.3061130, 9055436, Li_2021_ICCV, choi2009they, Volleyball}.
In these efforts, the primary challenge is to comprehend the spatiotemporal relationships between individuals in a scene~\cite{Volleyball}.

The Collective Activity Dataset~\cite{choi2009they} and the Volleyball Dataset~\cite{Volleyball} are two widely used video datasets for GAR. The former comprises more than 40 brief video clips including people crossing, waiting, queuing, walking, and talking. 
The Volleyball Dataset contains 1525 frames from 15 YouTube volleyball videos annotated for GAR purposes~\cite{Volleyball}.
In the Collective Activity Dataset, individuals are engaged in separate activities, but by virtue that others nearby are engaged in the same or similar activity, it
becomes easier to recognize an individual’s activity. This scenario aligns with the goal of this paper. Conversely, the Volleyball Dataset represents situations where a group of individuals may be carrying out different individual tasks in pursuit of the group's larger objective. The applicability of our approach to scenarios similar to this is discussed in Section~\ref{sec:discussion}.

\begin{table}[!t]
\small
\centering
\caption{Existing Datasets}
\label{tab:dataset}
\vspace{-10pt}
\scalebox{0.9}{
\begin{tabular}{|c|c|c|c|}
\hline
\textbf{Dataset} & \textbf{\# of Subject} & \textbf{\# of Activity}                                 & \textbf{Activity sample}                                                      \\ \hline
mHealth~\cite{banos2014mhealthdroid}& 10                     & 12                                                      & stand, sit, walk                                                                \\ \hline
Opportunity~\cite{Opportunity}& 12                     & \begin{tabular}[c]{@{}c@{}}5\\ (HighLevel)\end{tabular} & \begin{tabular}[c]{@{}c@{}}get up, break, clean,\\ sandwich, coffee\end{tabular} \\ \hline
KU-HAR~\cite{SIKDER202146}& 90                     & 18                                                      & stand, lay, run, walk                                                            \\ \hline
HAR~\cite{anguita2013public}& 30                     & 15                                                      & stand, lay, run, walk                                                            \\ \hline
MobiAct~\cite{vavoulas2016mobiact}& 67                     & 6                                                       & walk, stand, jog                                                                \\ \hline
MotionSense~\cite{motionsense}& 24                     & 6                                                       & walk, stand, jog                                                                \\ \hline
\end{tabular}}
\vspace{-1.5\baselineskip}
\end{table}

Several computer vision models have been proposed for GAR. Zhou et al. developed a Generative Model with high accuracy on the Collective Activity Dataset~\cite{10.5555/3061053.3061130}. Shu et al. proposed a graph LSTM-in-LSTM (GLIL) model that accomplished similarly high accuracy on the same dataset~\cite{9055436}. Li et al. introduced GroupFormer, a transformer-based model for GAR that achieved high accuracy on the Volleyball dataset~\cite{Li_2021_ICCV}. These results demonstrate that leveraging information about co-located individuals can significantly boost the accuracy of identifying the activity of an individual.

To date, techniques for GAR require a centralized view of the group performing the activity, exhibit privacy concerns associated with collecting and processing video, and incur the high overhead costs of computer vision. Nevertheless, these techniques do indicate some promising directions. For instance, work with the Collective Activity Recognition dataset implies the merit of using an individual's neighbors to corroborate an individual's activity.
To the best of our knowledge, there is no prior work that has gathered and studied group activity recognition (GAR) using IMU-based data.
In this paper, we explore this gap and the potential for using the activity of a group to corroborate the activity of an individual.

\section{Dataset and Experimental Setup}

To our knowledge, there is no  dataset that utilizes Inertial Measurement Units (IMU) to concurrently capture data from multiple participants engaged in the same activity. This dataset, the Group Work and Study (GWS) dataset, is driven by our research needs, offers support for gaining a deeper, nuanced understanding of group activities, and lays a groundwork for future investigations. Beyond this, it also sets a benchmark in the field, offering a blueprint for the generation of larger synthetic datasets in future studies.

To collect the measurements to construct the GWS dataset, we recruited 20 participants who were each provided a sensor to wear during their regular daily group activities. The participants were instructed to perform their activities without any additional restrictions. The duration of each data collection session varied depending on the activity, ranging between 5 minutes to 3 hours.

We utilized the Movesense Active sensor\footnote{\url{https://www.movesense.com/movesense-active/}}. The device was worn on the wrist of the participant's dominant hand. Prior to data collection, all sensors were swung together to ensure the synchronicity of sensor readings. Data collection from the sensors included information from the accelerometer and gyroscope at a frequency of 100 Hz. Due to its energy and memory limitations, data had to be collected synchronously and at short distances during the actvities. Despite our best efforts, the data stream from the sensor was still unstable, resulting in some missing data.

We collected 58.37 hours of IMU data from 11 data collection sessions. Detailed statistical information is presented in Table~\ref{tab:large_data}. The full dataset will be released publicly upon the publication of the paper. The activities captured in the dataset include:

\begin{enumerate}
    \item \textbf{Eating:} Multiple people having a meal together.
    \item \textbf{Lecture:} Multiple people attending a lecture.\footnote{The lecturer is included in the dataset but is omitted for our purposes in Section~\ref{sec:evaluation}.}
    \item \textbf{Meeting:} Multiple people in a meeting in the same room.
    \item \textbf{Office:} Multiple people working in a shared office space.
\end{enumerate}

\begin{table}[!t]
\caption{Statistics for the Group Work \& Study (GWS) Dataset}
\label{tab:large_data}
\small
\vspace{-10pt}
\begin{tabular}{|c|c|c|c|}
\hline
\textbf{Activity} & \textbf{\begin{tabular}[c]{@{}c@{}}Total Length \\ (Hours)\end{tabular}} & \textbf{\begin{tabular}[c]{@{}c@{}}Number of \\ Sessions\end{tabular}} & \textbf{\begin{tabular}[c]{@{}c@{}}Average Number of \\ Participants per Session\end{tabular}} \\ \hline
Eating            & 5.00022                                                                  & 2                                                                      & 3                                                                                              \\ \hline
Lecture           & 39.992534                                                                & 4                                                                      & 4.75                                                                                           \\ \hline
Meeting           & 9.850402                                                                 & 3                                                                      & 4                                                                                              \\ \hline
Office            & 3.525118                                                                 & 2                                                                      & 3                                                                                              \\ \hline
\textbf{Total}    & \textbf{58.368274}                                                       & \textbf{11}                                                            & \textbf{3.91}                                                                                  \\ \hline
\end{tabular}
\vspace{-\baselineskip}
\end{table}

The data collection occurred ``in the wild'', so participants are not intentionally limiting their movements to those germane to the purpose. For instance, during a lecture or meeting, participants may twist their hair or touch their faces. Some participants were observed working on other things during the lecture or sending a text message during the meeting. 
In addition, since the sensors were worn on the wrist like a watch, some participants were observed fidgeting with the sensor itself during data collection.
All of these are natural behaviors that would occur in a real setting.

\section{Corroborated Activity Recognition}\label{sec:system}

Human activity recognition (HAR) in complex settings can be challenging due to the variability and ambiguity of sensor data. We explore using information from the surroundings to improve the performance of HAR in complex settings. Specifically, we 
corroborate a local activity recognition result against activity recognition results collected from other nearby individuals to improve the accuracy and robustness of HAR. 

{\bf Local Recognition.}
To test such an approach, we use two different models as our backbone activity recognition models: Random Forest and DeepConvLSTM. Random Forest is a classic ensemble learning model that has been widely used in HAR~\cite{balli2019human, 9010115,8546311,THAKUR2022103417}, while DeepConvLSTM is a state-of-the-art deep learning model that has shown promising results when applied to HAR~\cite{s16010115}.
Each device runs one of these state-of-the-art backbone models locally. This model takes as input the locally sensed IMU data and generates a set of probabilities that indicate the likelihood that the local IMU data corresponds to each of a set of different activities. Ordinarily, each device would then use its local results to independently identify the most likely activity. We employ a sliding window approach, where each device continuously collects raw data, but generates a set of probabilities for the dictionary of activities every 5 seconds, using the raw data from the previous 10 seconds. These settings were based on empirical observations to achieve a balance of responsiveness and overhead, but of course, these are tunable parameters for specific application deployments. They could also adjust dynamically in response to changing conditions, e.g., rapidly changing activity.

{\bf Activity Sharing.}
Each device 
periodically broadcasts its computed probabilities to other nearby clients over lightweight device-to-device communication~\cite{Bluetooth5, WiFiDirect}. In our prototype system, the updated probabilities are shared immediately (i.e., every five seconds) with neighboring devices. In practice, the broadcast frequency could be tuned according to the rate at which we expect individuals' activity to change, the rate at which we expect a device's neighbor set to change, concerns about energy consumption of communication and computation, or some combination of these factors. By sharing the probabilities rather than the raw data, we can reduce the amount of data that needs to be transmitted and lower the computational load on the devices because they only perform activity recognition on their own raw data.  

\begin{figure}[!t]
  \centering
  \includegraphics[width=.4\textwidth]{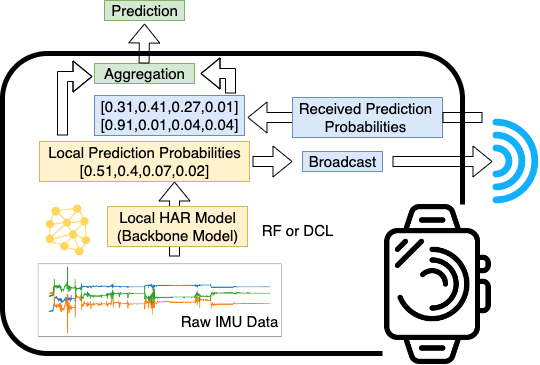}
  \vspace{-\baselineskip}
  \caption{System Diagram}
  \label{fig:system_diagram} 
  \vspace{-\baselineskip}
\end{figure}

An additional important benefit of this approach is that it allows each device to use a machine-learning model that is best suited to its computational resources or appropriate for its particular sensing data. By allowing different devices to select different models, we can optimize the trade-off between model accuracy and computational complexity. For example, a high-end device with ample computational resources can use a more complex model that achieves higher accuracy, while a low-end device with limited computational resources can use a simpler model that still achieves reasonable accuracy. Instead of synchronizing on a specific model architecture, the system has devices synchronize on the output representation of a set of activity probabilities. This flexibility enables us to achieve real-time activity recognition across a range of devices in a distributed setting with varying capabilities and constraints. Figure~\ref{fig:system_diagram} shows a system diagram.

{\bf Aggregating Information.} Rather than immediately confirming the locally recognized activity, every device continuously collects activity probabilities from any neighbors in the surroundings. 
Each time the device generates a new set of local probabilities (i.e., every five seconds in our prototype), we combine the local  probabilities with the most recent received from each neighboring device. In our prototype, we assume perfect communication (no loss), and therefore each aggregation simply includes the neighbor probabilities received over the past five seconds.
We  compute the average probability for each activity in the activity set using a simple per-activity mean across all received samples, including the probabilities from the local device.
This is a simple aggregation scheme for our proof-of-concept, though alternative approaches could also be explored, for instance using majority voting or computing a mean that more heavily weights the local measurements in contrast to the neighbors' activity information. %

\begin{figure*}[!t]
    \centering
    \begin{subfigure}{0.24\textwidth}
        \centering
        \includegraphics[width=\textwidth]{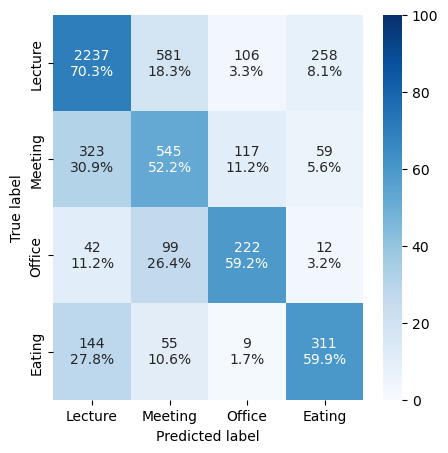}
        \caption{DCL without Corroboration}
    \end{subfigure}\hfill
    \begin{subfigure}{0.24\textwidth}
        \centering
        \includegraphics[width=\textwidth]{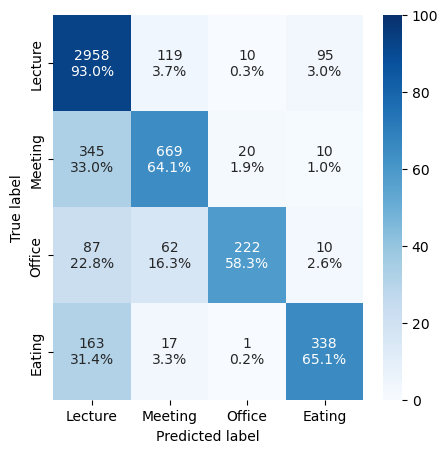}
        \caption{RF without Corroboration}
    \end{subfigure}\hfill
    \begin{subfigure}{0.24\textwidth}
        \centering
        \includegraphics[width=\textwidth]{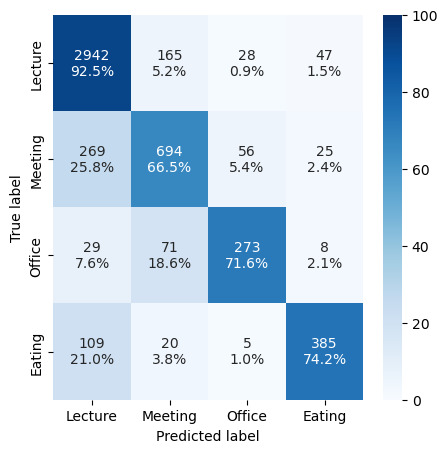}
        \caption{DCL with Corroboration}
    \end{subfigure}\hfill
    \begin{subfigure}{0.24\textwidth}
        \centering
        \includegraphics[width=\textwidth]{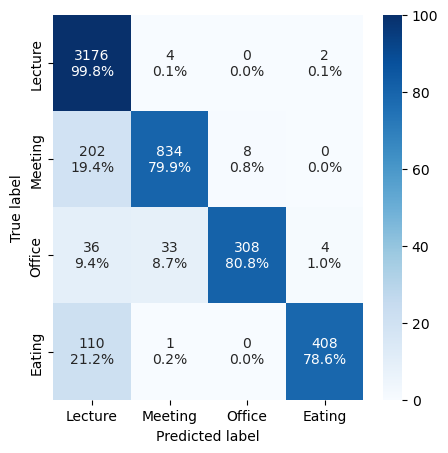}
        \caption{RF with Corroboration}
    \end{subfigure}
    \caption{Confusion Matrix for Group Work and Study (GWS) Dataset}
    \label{fig:BigDatasetResult}
\end{figure*}

\section{A Prototype Evaluation}\label{sec:evaluation}
To evaluate our approach, we implemented the system described above using two backbone HAR models: a Random Forest classifier (RF) and DeepConvLSTM (DCL). For each, we conducted a hyperparameter search to identify the combination of hyperparameters that achieved the best balance between accuracy and computational efficiency. For DCL, we found that a network architecture with 2 convolutional layers, each with 64 filters of size 5, connected to 1 layer of LSTM with 128 hidden nodes, and ending with a fully connected layer, provided the best performance. We chose these hyperparameters based on their ability to effectively capture the temporal and spatial dynamics of the sensor data. For RF, we used 100 decision trees in the forest, with the gini criterion and a minimum number of samples required to split an internal node of 2.

As described above, we apply a sliding window approach to process the raw IMU data. In the case of DCL, we feed in the raw IMU data and obtain the resulting probabilities for each activity. In the case of the RF, we first extract 
features such as the mean and variance from each window\footnote{We use the following features in this paper: mean, variance, maximum, minimum, skewness, kurtosis, total energy, signal magnitude area (SMA), and zero-crossing rate.} and feed these features into the RF classifier for human activity recognition.

For testing, we randomly selected $20\%$ of the windows for testing and used the remaining windows as the data used to train. The dataset is imbalanced, so SMOTE technique is applied to balance the training set before training the model. SMOTE creates synthetic samples of the minority class by interpolating between existing samples~\cite{Chawla_2002}. This approach helped to mitigate the class imbalance and improve the model's ability to generalize to new data.

\begin{table}[!t]
\centering
\caption{Results for both with and w/o corroboration (Corr.)}
\label{tab:large_result}
\small
\vspace{-10pt}
\begin{tabular}{|c|c|c|c|c|}
\hline
\textbf{}         & DCL & RF & \begin{tabular}[c]{@{}c@{}}DCL (Corr.)\end{tabular} & \begin{tabular}[c]{@{}c@{}}RF (Corr.)\end{tabular} \\ \hline
Accuracy          & 0.6475       & 0.8168        & 0.8377                                                         & \textbf{0.9220}                                                   \\ \hline
Recall (Macro)    & 0.6041       & 0.7011        & 0.7619                                                         & \textbf{0.8479}                                                  \\ \hline
Precision (Macro) & 0.5538       & 0.8069        & 0.7978                                                         & \textbf{0.9545}                                                  \\ \hline
\end{tabular}
\vspace{-\baselineskip}
\end{table}

The Group Work and Study (GWS) Dataset includes four activities: attending a lecture, participating in an in-person meeting, working in a shared office space, and eating in a group. Table~\ref{tab:large_result} and Figure~\ref{fig:BigDatasetResult}
show the accuracy of the activity recognition models for the GWS Dataset, both with and without using corroborating information shared by the devices of other nearby individuals.

Comparing the performance of the DCL and RF models, we found that RF achieved higher accuracy for this dataset. More importantly for our contribution, however, when comparing the same backbone model with and without corroborating information from neighbors, we observed significant improvements in accuracy for both models. Specifically, for DCL, the corroborating information improved accuracy by $19.02\%$, while for RF, it improved accuracy by $10.52\%$. These results demonstrate the potential of neighborhood corroboration in improving the accuracy and robustness of machine-learning models applied in complex settings. From Figure~\ref{fig:BigDatasetResult}, as expected, we can observe that the model struggles to differentiate between activities such as listening to a lecture, having a meeting, and working in an office when the information from others nearby is not used.

\section{Conclusion and Future Work}

\label{sec:discussion}
We explored a new approach to activity recognition by leveraging information from nearby individuals to corroborate a local prediction. We collected a novel dataset and tested our approach using two models commonly applied to HAR: DeepConvLSTM and Random Forest. Our results demonstrated that corroboration with the activity of others nearby can significantly improve activity recognition accuracy, with a minimum improvement of $10.52\%$. These results suggest that activity recognition with corroboration has the potential to enable more robust and accurate machine-learning models in complex settings. Moreover, our approach is computationally efficient and requires minimal data transmission, making it well-suited for resource-constrained devices and distributed systems.

In the future, we plan to expand upon this work by collecting more data from individuals performing even more diverse sets of activities. In the introduction, we scoped this first effort to focus exclusively on group activities where all of the individuals in the activity are expected to have the same or very similar ``low-level'' activities recognized by the underlying HAR scheme. However, we also introduced the potential for our approach to be extended to recognize group activities comprising individuals engaged in different low-level activities that, when combined, generate a high level situation (e.g., players engaged in a team sport). The basic approach described here provides a foundational step in addressing this more complex problem, where the novel insights required are in how to aggregate diverse prediction probabilities in to a higher level situation prediction.

Undoubtedly, the substantial increase in activity recognition accuracy we achieved by leveraging nearby individual activities opens up new possibilities for the development of highly effective and efficient machine learning system for human activity recognition.

\newpage
\bibliographystyle{ACM-Reference-Format}
\bibliography{references.bib}
  
\end{document}